\pdfoutput=1
\documentclass[letterpaper]{article}
\usepackage[preprint]{aaai2027}
\usepackage[hyphens]{url}
\usepackage{graphicx}
\usepackage{natbib}
\usepackage{caption}
\usepackage{booktabs}
\usepackage{amsmath}
\usepackage{amssymb}
\usepackage{xspace}
\usepackage{array}
\usepackage{multirow}
\usepackage{colortbl}
\usepackage{placeins}
\usepackage{float}

\newcommand{\hd}[1]{\textbf{#1}}
\newcommand{\best}[1]{\textbf{#1}}

\newcommand{\fk}[1]{#1}
\newcommand{\pp}[1]{#1}


\newcommand{\sysname}{XEWorld\xspace}

\title{\sysname: Can Action-Conditioned World Models \\ Generalize to Unseen Robot Embodiments?}
\author{
    Yixiang Chen\textsuperscript{\rm 1,\rm 2},
    Jiabing Yang\textsuperscript{\rm 1,\rm 2},
    Yuan Xu\textsuperscript{\rm 1,\rm 2},
    Qisen Ma\textsuperscript{\rm 1,\rm 2},
    Keji He\textsuperscript{\rm 3},
    Peiyan Li\textsuperscript{\rm 1,\rm 2},
    Kai Wang\textsuperscript{\rm 1,\rm 2},\\
    Ziheng He\textsuperscript{\rm 1,\rm 2},
    Xiangnan Wu\textsuperscript{\rm 1,\rm 2},
    Jing Liu\textsuperscript{\rm 4},
    Nianfeng Liu\textsuperscript{\rm 4},
    Yan Huang\textsuperscript{\rm 1,\rm 2,\rm 4}\setcounter{footnote}{1}\thanks{Corresponding authors.},
    Liang Wang\textsuperscript{\rm 1,\rm 2}\footnotemark[2]
}
\affiliations{
    \textsuperscript{\rm 1}New Laboratory of Pattern Recognition (NLPR), Institute of Automation, Chinese Academy of Sciences\\
    \textsuperscript{\rm 2}School of Artificial Intelligence, University of Chinese Academy of Sciences\\
    \textsuperscript{\rm 3}Shandong University \qquad
    \textsuperscript{\rm 4}FiveAges\\
    yixiang.chen@cripac.ia.ac.cn, \{yhuang, wangliang\}@nlpr.ia.ac.cn
}

\begin{document}

\maketitle

\begin{abstract}
Action-conditioned world models are promising learned simulators for robotic manipulation, yet evaluating them exclusively on training robots fails to reveal whether they capture physical dynamics or merely memorize visual patterns. To answer whether a model can faithfully render a robot it has never seen, we introduce \textbf{\sysname{}}, a controlled cross-embodiment testbed for world models that isolates embodiments by evaluating held-out robots within physically identical scenes. Our systematic analysis uncovers a shared architectural bottleneck: current models act primarily as 2D visual pattern matchers whose generalization is governed by visual similarity rather than physical kinematic similarity. Driven by this limitation, they struggle to translate abstract numeric joint actions into coherent visual trajectories, and fail to predict dynamic visual changes from static initial observations. Consequently, successfully rendering an unseen embodiment zero-shot strictly requires heavily grounded cues, specifically pixel-space actions and explicit spatial-temporal alignment. Even when bypassing this zero-shot barrier via few-shot adaptation, the forced appearance recovery triggers catastrophic forgetting of seen embodiments. Together, these failures expose a critical inability to apply learned physical dynamics to novel visual appearances, highlighting that achieving true cross-embodiment generalization requires architectural innovations that decouple visual appearance from underlying physical dynamics.
\end{abstract}

\begin{figure*}[t]
\centering
\includegraphics[width=0.7\textwidth]{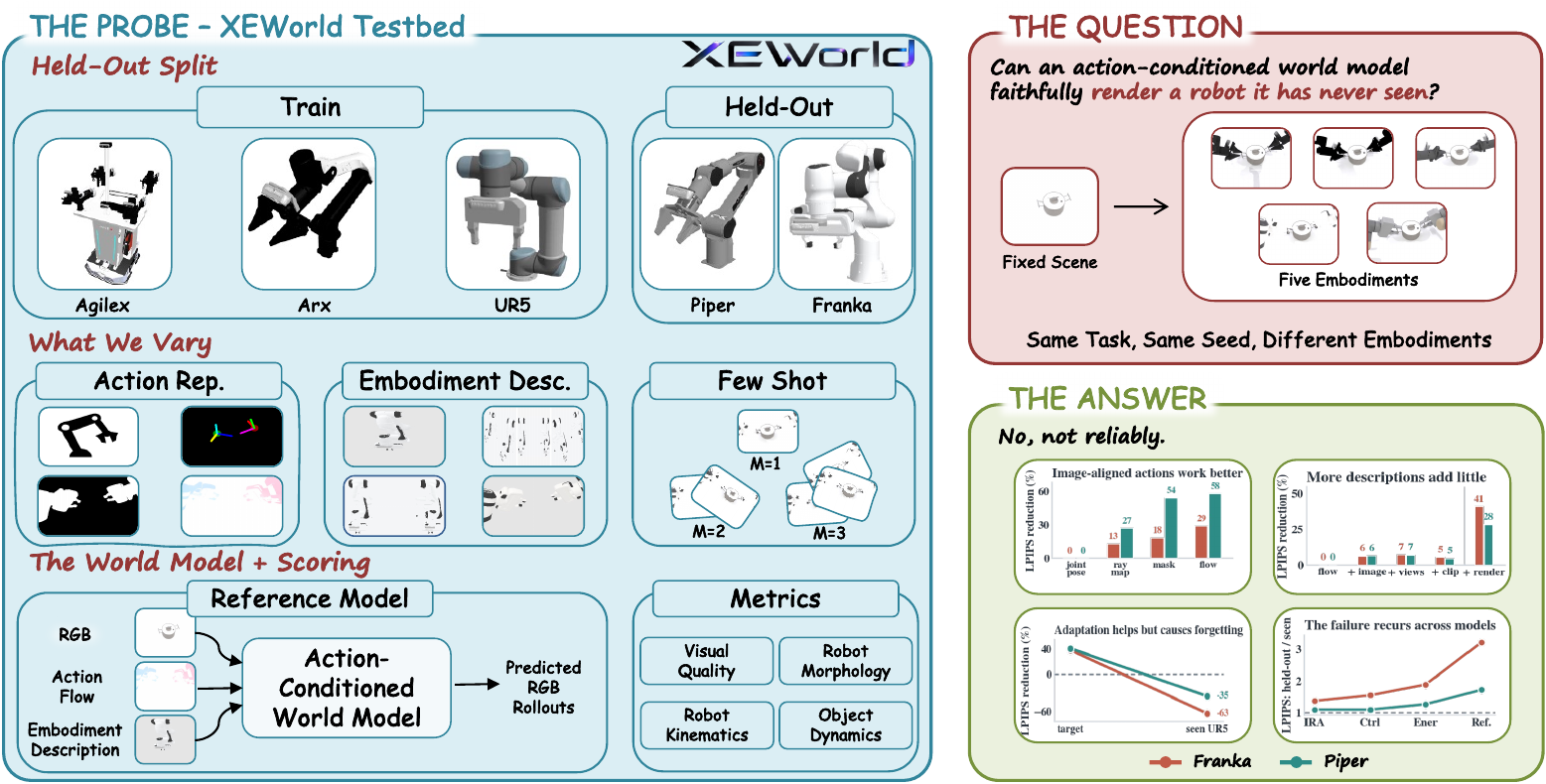}
\caption{\textbf{The \sysname{} cross-embodiment testbed.} We introduce XEWorld to investigate whether action-conditioned world models genuinely capture generalizable physical dynamics, or merely memorize embodiment-specific visual patterns.}
\label{fig:setup}
\end{figure*}

\section{Introduction}

Action-conditioned world models predict the visual consequences of an embodiment's actions over time.
These learned simulators hold great potential for downstream tasks such as offline policy evaluation and planning \citep{unisim,irasim,cosmos,dreamgen}.
To serve as reliable simulators, however, these models must capture invariant physical dynamics rather than memorizing embodiment-specific visual patterns.
Yet, current testbeds evaluate world models exclusively on their training embodiments.
Such in-distribution evaluations cannot distinguish a model that genuinely captures the underlying physical dynamics from one that merely memorizes visual appearances, as both perform equally well on seen robots.

To truly verify if world models learn generalizable action-conditioning, it is crucial to assess their cross-embodiment capabilities.
This requirement raises a fundamental question: \emph{can an action-conditioned world model faithfully render a robot it has never seen?}
While the surrounding scene and object dynamics remain constant, the robot's physical structure varies.
By isolating the embodiment as the primary variable, cross-embodiment prediction becomes a powerful probe to test if a world model can decouple the physical environment from the specific embodiment.
To this end, we introduce \textbf{\sysname}, a systematic testbed enabling controlled empirical studies of cross-embodiment generalization.

Specifically, \sysname{} achieves this strict control by evaluating models on held-out robots performing the same 25 manipulation tasks within physically identical scenes.
Leveraging this strict control, we can rigorously re-examine two widespread assumptions about cross-embodiment transfer: (1) that generalization naturally succeeds on embodiments with similar kinematics, and (2) that visual mismatches on unseen robots can be resolved simply by supplying static visual references of the target embodiment.
Our analysis reveals that both expectations face critical limitations, exposing a fundamental disconnect between matching a static visual pattern and simulating dynamic physical motion.

Our empirical results demonstrate that a model's capacity to transfer to a novel embodiment is governed primarily by visual similarity rather than physical kinematic similarity.
Consequently, current architectures behave more like 2D visual pattern matchers than systems that genuinely understand physical dynamics.
This fundamental nature explains why they struggle severely to render visually coherent videos from abstract numeric joint actions, forcing a reliance on pixel-space actions to bypass this translation gap.
This conclusion is further substantiated by our distance analysis: generation error correlates strongly with the visual appearance distance from the training fleet, yet shows a weak and unstable relationship with physical reachable-workspace distance.

Crucially, this 2D pattern-matching bottleneck persists even when the model receives explicit target descriptions.
Despite the prevailing assumption that rich visual cues, such as multi-view images or embodiment-specific motion clips, can resolve appearance failures, we find such unaligned inputs yield negligible gains.
Lacking structural awareness, the model cannot dynamically register static visual features to dynamic action inputs.
Consequently, successfully rendering an unseen embodiment zero-shot requires explicit spatial-temporal alignment, such as perfectly registered per-frame renderings.
Alternatively, directly fine-tuning the model on target demonstrations can bypass this alignment barrier and force appearance recovery.
However, this few-shot adaptation triggers catastrophic forgetting of seen embodiments.
This reveals that current architectures cannot dynamically bind a novel appearance to motion without overwriting their learned priors.

Our main contributions are threefold.

\begin{itemize}
\item We introduce \textbf{\sysname{}}, a controlled cross-embodiment testbed featuring physically identical scenes across five robot embodiments. By isolating the embodiment as the primary variable, it enables systematic evaluations of how current architectures decouple physical environments from specific embodiments.
\item We conduct a systematic empirical study on cross-embodiment generalization of world models. Through rigorous distance analysis, we demonstrate that generation quality is governed by visual similarity rather than physical kinematic similarity, revealing that current architectures act primarily as 2D visual pattern matchers.
\item We evaluate a series of controlled interventions spanning action representations, static visual descriptions, and few-shot fine-tuning. We find that explicit spatial-temporal alignment and pixel-space actions are fundamental prerequisites for reliable transfer, providing concrete design guidelines for future world models.
\end{itemize}

\section{Related Work}
\label{sec:related}

\paragraph{Action-conditioned embodied world models for robotics.}
Action-conditioned world models have emerged as learned simulators for policy evaluation, data generation, and planning \citep{unisim,ivideogpt,irasim,cosmos,dreamgen,genie,qwenrobotworld,bridgev2w,kinema4d,maskedvisualactions}.
By predicting future video frames conditioned on robot actions, these systems can capture complex object interactions and scene dynamics.
While recent works provide models for fine-grained manipulation rollouts \citep{irasim}, physical AI \citep{cosmos,cosmos3}, and trajectory synthesis \citep{dreamgen,egodemogen}, they are typically evaluated on the same robot.
In contrast, \sysname{} explicitly evaluates world models on held-out embodiments, shifting the focus from rendering quality on familiar platforms to physical generalization on unseen embodiments.

\paragraph{Benchmarks for video world models.}
Standard video benchmarks \citep{vbench,worldscore,meng2024towards} focus primarily on perceptual quality, lacking action-conditioning or physical verification.
Recent benchmarks like WorldArena \citep{worldarena,worldarena2} evaluate functional utility, while DreamGen Bench \citep{dreamgen} evaluates robot adaptation using VLM-based judgments.
However, they measure performance on seen target embodiments and lack kinematics-grounded metrics or counterfactual isolation.
Without explicit physical verification, it remains unclear whether the generated videos truly respect the geometric and mechanical constraints of the robot.
To address this, \sysname{} serves as a diagnostic testbed that holds the embodiment out, isolates it with paired counterfactual data, and rigorously scores joint motion against forward-kinematics ground truth.

\paragraph{Cross-embodiment robot learning.}
Generalizing across different physical structures is a long-standing goal in cross-embodiment robot learning \citep{anybody,embodimentscaling}.
Deploying a single control policy across multiple platforms typically suffers from severe visual and kinematic domain shifts.
Recent efforts aggregate diverse datasets \citep{openx,oxeauge} and multi-embodiment simulators \citep{robotwin,robotwin2} to improve cross-embodiment policy generalization.
While this literature \citep{beingh05} primarily investigates whether a policy can generalize across different robot bodies, our work shifts the focus to whether an action-conditioned world model can do so.
These two directions are highly complementary: a generalizable world model can provide the counterfactual rollouts necessary for evaluating cross-embodiment policies.

\section{The \sysname Testbed}

\sysname{} is a controlled testbed for evaluating cross-embodiment generalization in action-conditioned world models, designed to isolate the robot's physical body as the sole experimental variable.
It is organized around a central research question:
\emph{Can an action-conditioned world model faithfully render a robot it has
never seen?}
The remainder of this section outlines the construction of \sysname{}. We first introduce the held-out-embodiment protocol for enabling controlled cross-embodiment evaluations, then define a decoupled metric suite to separate morphology from kinematics, and finally formalize embodiment distances to quantify generalization difficulty.

\subsection{Held-Out-Embodiment Protocol}
\label{sec:protocol}

\paragraph{Embodiments.}
The testbed uses five bimanual robot setups re-rendered on the RoboTwin simulator \citep{robotwin2}.
It supports flexible data splits, primarily utilizing a fixed 3-train / 2-held-out split for intervention studies and a leave-one-embodiment-out (LOEO) split for distance analysis.
In the main 3-train / 2-held-out split, the training robots are Aloha-Agilex, Arx-X5, and UR5.
The held-out pair is chosen to separate two confounded factors.
We evaluate unseen-robot generalization on two distinct held-out embodiments: Franka Panda and Piper.
While their physical kinematic distances to the training fleet are similar, their visual appearance distances differ significantly.
We report their results separately and never average them, ensuring that performance drops can be accurately isolated and analyzed across different levels of visual similarity.

\paragraph{Paired cross-embodiment data.}
The testbed's core asset is its strictly paired cross-embodiment data.
For a given task and seed, the scene layout, object poses, lighting, and camera are byte-identical across all five robots, so the only varying factor is the body.
This byte-level alignment guarantees that any performance drop on an unseen robot stems entirely from its embodiment.
The release covers 25 manipulation tasks $\times$ 5 robots at a camera resolution of $320\times240$, with 50 paired training seeds per task and a disjoint set of 20 held-out test seeds per task.
Each episode stores RGB images, joint actions, rigid and articulated object poses, contact states, and camera parameters.
These rich state annotations provide the exact ground truth required by our decoupled metric suite.

\paragraph{Evaluation protocols.}
Models are evaluated on their ability to predict future video frames given an initial state and an action sequence.
To systematically diagnose generalization bottlenecks, \sysname supports two main evaluation settings.
First, the zero-shot setting tests how well models can generalize to a new embodiment given only its descriptive specifications (e.g., a reference image or its URDF geometry), without any training data.
Second, the few-shot setting evaluates the model's adaptation capability by providing real demonstrations of the held-out robot.

\subsection{Decoupled Metric Suite}
\label{sec:metrics}

\sysname grades a predicted rollout along four distinct visual and physical dimensions.
\begin{itemize}
\item \textbf{Visual Quality:} SSIM, PSNR, and LPIPS \citep{lpips} against the ground-truth video. This captures standard image-level fidelity, which is the primary focus of existing video generation benchmarks.
\item \textbf{Robot Morphology:} Symmetric IoU, boundary $F_1$, and region-LPIPS between the predicted and ground-truth robot masks (segmented via SAM2 \citep{sam2}). This evaluates whether the model renders the correct physical shape and appearance of the new embodiment.
\item \textbf{Robot Kinematics:} Keypoint reprojection accuracy (Percentage of Correct Keypoints, PCK, reported at normalized tolerance $\alpha{=}0.1$ throughout), keypoint pixel error, and normalized DTW, comparing URDF-derived forward kinematics against predictions tracked by CoTracker3 \citep{cotracker3}. This measures whether the generated robot moves according to its correct joint constraints.
\item \textbf{Object Dynamics:} Object-trajectory pixel error against the simulator's recorded object poses, tracked with CoTracker3 \citep{cotracker3}. This assesses whether the robot's physical interactions move the objects correctly.
\end{itemize}
We report all evaluations across these four dimensions to provide a comprehensive, multi-faceted diagnostic of cross-embodiment generalization. Rather than collapsing these distinct physical and visual aspects into a single scalar, we analyze them individually to pinpoint exactly where a model succeeds or fails.

\begin{table}[t]
    \centering
    \small
    \setlength{\tabcolsep}{6pt}
    \renewcommand{\arraystretch}{1.1}
    \begin{tabular}{@{}lccc@{}}
    \toprule
    \multicolumn{1}{c}{\hd{Robot}} & \hd{Role} & \hd{Kin.\ Dist.} & \hd{App.\ Dist.} \\
    \midrule
    Aloha-Agilex & seen & 0.144 & 0.175 \\
    Arx-X5       & seen & 0.095 & 0.063 \\
    UR5          & seen & 0.070 & 0.058 \\
    \midrule
    \fk{Franka} & \fk{held-out, far}  & 0.082 & 0.107 \\
    \pp{Piper}  & \pp{held-out, near} & 0.074 & 0.052 \\
    \bottomrule
    \end{tabular}
    \caption{\textbf{Distance of each robot to the rest of the fleet.}}
    \label{tab:dist}
\end{table}

\begin{figure*}[t]
    \centering
    \includegraphics[width=0.85\textwidth]{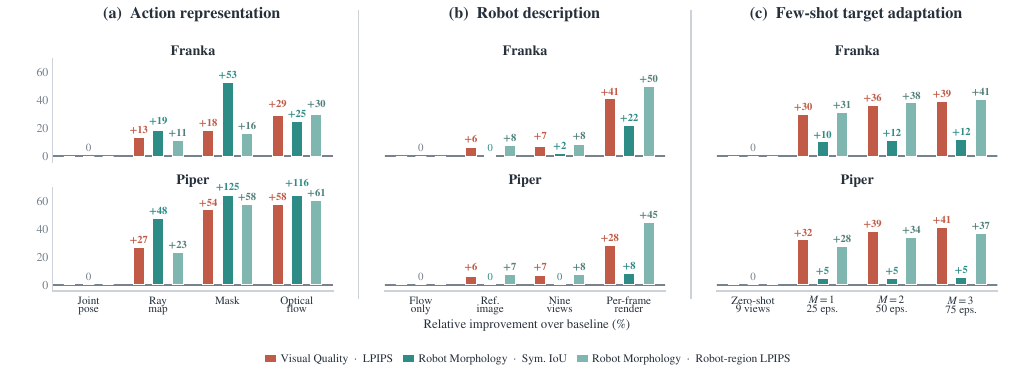}
    \caption{\textbf{Impact of targeted interventions on unseen-embodiment generation.}
    Bars report relative improvements over the baseline for the two held-out robots across different intervention strategies.}
    \label{fig:money}
\end{figure*}

\subsection{Embodiment Distance}
\label{sec:distance}

To investigate what makes a held-out robot difficult to generalize to, we quantify each robot's distance to the rest of the training fleet along two axes.
The \textbf{kinematic distance} is computed as the workspace Chamfer distance between point clouds sampled from each robot's forward-kinematics reachable set. It captures the structural differences in how the bodies move and reach.
The \textbf{appearance distance} is measured as the cosine distance between visual features extracted from each robot's standardized reference renders, capturing differences in color distribution and projected silhouette.
For each robot, we report its distance to the other four (Table~\ref{tab:dist}).

\section{Reference World Model}
\label{sec:model}

To systematically study cross-embodiment generalization, we need a model that explicitly decouples a robot's motion from its visual appearance.
We adapt FlowWAM \citep{flowwam}, an action-conditioned embodied world model, and fine-tune it on our three training robots.
We design the model to process three distinct input streams: (1)~scene RGB for environmental context, (2)~a pixel-space action signal, and (3)~an embodiment specification stream.
The action stream is robot-masked optical flow, computed with RAFT \citep{raft} on a robot-only forward-kinematics render of the commanded action sequence rather than on the ground-truth scene video.
The specification stream acts as a visual prompt (e.g., images or videos of the robot) that tells the model what the current embodiment looks like.
By isolating morphology into this separate stream, the model learns to follow the optical flow for motion dynamics while cross-attending to the specification to render the correct body.

Crucially, this decoupled architecture allows us to control exactly how much morphological information of the target embodiment the model receives during inference.
When evaluating on a novel embodiment, we keep the optical flow fixed and vary the richness of the specification stream.
This naturally forms an intervention ladder with five levels of specification richness: (0)~\emph{flow only} (an empty specification stream), (1)~\emph{+reference image} (a single front-view render), (2)~\emph{+multi-view} (nine renders from different angles), (3)~\emph{+articulation clip} (a short video of the robot sweeping all its joints), and (4)~\emph{per-frame robot render} (the target embodiment rendered with forward kinematics at every time step).

\section{Experimental Results}

Our experiments aim to systematically examine the causes of cross-embodiment generalization failure. We structure our evaluation to isolate these bottlenecks in two stages. First, using a controlled reference model, we quantify the baseline generalization gap and evaluate whether this gap can be reduced by altering the action representation, providing richer target-robot descriptions, or applying few-shot adaptation. Figure~\ref{fig:money} summarizes the relative effect of each intervention protocol. Second, we investigate whether embodiment distance predicts this generalization difficulty, and finally examine whether the same degradation exists across multiple current world model architectures.

\begin{table*}[t]
\centering
\footnotesize
\setlength{\tabcolsep}{3.2pt}
\renewcommand{\arraystretch}{1.14}
\begin{tabular*}{0.84\textwidth}{@{\extracolsep{\fill}}cc cc cc cc c@{}}
\toprule
\multirow{3}{*}{\shortstack[c]{\hd{Action}\\\hd{representation}}} &
\multirow{3}{*}{\hd{Target}} &
\multicolumn{2}{c}{\hd{Visual Quality}} &
\multicolumn{2}{c}{\hd{Robot Morphology}} &
\multicolumn{2}{c}{\hd{Robot Kinematics}} &
\hd{Object Dynamics} \\
\cmidrule(lr){3-4}\cmidrule(lr){5-6}\cmidrule(lr){7-8}\cmidrule(l){9-9}
& & \multirow{2}{*}{PSNR\,$\uparrow$} & \multirow{2}{*}{LPIPS\,$\downarrow$} &
\multirow{2}{*}{Sym.\ IoU\,$\uparrow$} & \multirow{2}{*}{Boundary $F_1$\,$\uparrow$} &
\multirow{2}{*}{PCK\,$\uparrow$} & Keypoint px. & Trajectory px. \\
& & & & & & & error\,$\downarrow$ & error\,$\downarrow$ \\
\midrule
\multirow{2}{*}{Joint pose} & Franka & 17.4 & 0.254 & 0.534 & 0.306 & 0.280 & 32.4 & 15.4 \\
                            & Piper  & 18.6 & 0.228 & 0.358 & 0.208 & 0.219 & 39.1 & 15.4 \\
\addlinespace[1pt]
\multirow{2}{*}{Ray map} & Franka & 18.2 & 0.221 & 0.634 & 0.379 & 0.456 & 17.7 & \best{14.3} \\
                         & Piper  & 21.0 & 0.167 & 0.530 & 0.328 & 0.584 & 12.5 & 10.8 \\
\addlinespace[1pt]
\multirow{2}{*}{Mask} & Franka & 18.4 & 0.207 & \best{0.818} & \best{0.636} & \best{0.492} & \best{16.7} & 16.2 \\
                      & Piper  & \best{24.5} & 0.105 & \best{0.807} & \best{0.721} & \best{0.633} & \best{12.0} & \best{8.2} \\
\addlinespace[1pt]
\multirow{2}{*}{Optical flow} & Franka & \best{19.2} & \best{0.180} & 0.667 & 0.463 & 0.456 & 21.6 & 16.1 \\
                              & Piper  & 24.2 & \best{0.097} & 0.773 & 0.659 & 0.593 & 12.5 & 8.7 \\
\bottomrule
\end{tabular*}
\caption{\textbf{Impact of action representation on unseen-embodiment generation.}
Abstract numeric joint poses struggle to render novel morphologies. In contrast, pixel-space inputs (ray maps, masks, and optical flow) successfully bridge the spatial translation gap, yielding large observed improvements across visual quality and morphological fidelity.}
\label{tab:actrep}
\end{table*}

\subsection{How Does Action Representation Affect Visual Generalization?}
\label{sec:exp-actrep}

The action representation fundamentally constrains the model's capacity for cross-embodiment transfer. While numeric joint poses are exact, their spatial mapping is highly embodiment-specific, meaning a specific angle set corresponds to entirely different visual configurations across robots. In contrast, pixel-space signals directly anchor motion to the image plane. We compare four representations in Table~\ref{tab:actrep}: numeric joint poses, projected ray maps, per-frame robot masks, and robot-masked optical flow. All four are deterministic functions of the commanded action sequence, the target URDF, and the camera parameters, so none of them requires privileged access to the target rollout, but they differ in how much geometry they retain and in how that geometry is encoded for the network.

Our empirical results reveal a substantial advantage for pixel-space action inputs over numeric joint poses. Replacing abstract joint poses with optical flow reduces LPIPS error by 29\% on Franka and 57\% on Piper, while improving symmetric IoU by 25\% and 116\%, respectively. This indicates a profound representational bottleneck: current visual backbones struggle to decode abstract numeric kinematics into coherent spatial transformations for unseen morphologies. Pixel-space action inputs bypass this translation gap by sharing the geometric coordinate space of the visual output. Even sparse spatial projections, such as ray maps, recover a substantial portion of the lost performance, confirming that shifting from pure numeric values to spatially grounded signals is a critical prerequisite for transfer.

Comparing the dense spatial inputs (binary masks and optical flow) reveals a structural trade-off between shape preservation and perceptual realism. The mask condition yields the highest IoU and boundary $F_1$ scores, as it explicitly provides the structural silhouette at every timestep, forcing the generation process to respect rigid boundaries. Conversely, optical flow achieves superior overall perceptual quality (lower LPIPS). By guiding motion without strictly dictating boundary occupancy, optical flow grants the generative prior more flexibility to synthesize natural lighting and textures.

This divergence reveals a core design trade-off: an interface that rigidly dictates occupancy best preserves morphological boundaries, whereas a motion-field interface yields higher overall visual fidelity. Figure~\ref{fig:qual-actrep-cond}(a) corroborates this quantitatively observed trend. While pixel-space actions maintain structural coherence during movement, numeric joints frequently cause the unseen robot to dissolve or severely warp during object interaction.

\begin{table}[t]
    \centering
    \scriptsize
    \setlength{\tabcolsep}{3.1pt}
    \renewcommand{\arraystretch}{0.98}
    \begin{tabular}{@{}l ccc@{}}
    \toprule
    \multicolumn{1}{c}{\multirow{2}{*}[-0.35ex]{\hd{Conditioning}}} &
    \multirow{2}{*}[-0.35ex]{\hd{Sym.\ IoU\,$\uparrow$}} &
    \hd{Robot-region} & \hd{Full-frame} \\
    & & \hd{LPIPS\,$\downarrow$} & \hd{LPIPS\,$\downarrow$} \\
    \midrule
    \rowcolor{black!7}
    \multicolumn{4}{l}{\textit{\textbf{Franka}}} \\
    flow only        & 0.667 & 0.251 & 0.180 \\
    $+$ ref.\ image  & 0.667 & 0.231 & 0.168 \\
    $+$ nine views   & 0.679 & 0.230 & 0.167 \\
    $+$ artic.\ clip & 0.671 & 0.235 & 0.170 \\
    $+$ per-frame render & \best{0.815} & \best{0.126} & \best{0.106} \\
    \rowcolor{black!7}
    \multicolumn{4}{l}{\textit{\textbf{Piper}}} \\
    flow only        & 0.773 & 0.138 & 0.097 \\
    $+$ ref.\ image  & 0.775 & 0.128 & 0.090 \\
    $+$ nine views   & 0.777 & 0.128 & 0.090 \\
    $+$ artic.\ clip & 0.772 & 0.130 & 0.092 \\
    $+$ per-frame render & \best{0.836} & \best{0.076} & \best{0.069} \\
    \bottomrule
    \end{tabular}
    \caption{\textbf{Effect of richer static descriptions vs. per-frame rendering.}
    Adding unaligned visual cues provides marginal gains, whereas supplying a perfectly aligned per-frame render drastically improves morphological fidelity.}
    \label{tab:cond}
\end{table}

\subsection{Are Richer Robot Descriptions Sufficient?}
\label{sec:exp-cond}

A common assumption is that models fail on new robots simply because they do not know what the new robot looks like. We test this hypothesis empirically by feeding the model progressively richer descriptions of the unseen robot. We add a single reference image, then expand it to nine static views, and finally provide a dynamic articulation video clip. 

Table~\ref{tab:cond} shows that the benefit of these rich descriptions saturates almost immediately. The first reference image does help, cutting robot-region LPIPS by 8\% on Franka and 7\% on Piper, but the gain stops there. Moving from that single image to nine diverse views reduces the full-frame LPIPS by less than 1\% and increases shape IoU by less than 2\%. Replacing the static views with an articulation clip actually reverses some of these minor gains, likely because the unaligned motion in the clip distracts the generation process. The model therefore extracts a small, one-shot appearance benefit and then stops improving no matter how much more it is shown, which suggests that the bottleneck is not a lack of visual details about the target robot. 

Instead, the limiting factor is temporal and spatial alignment. The generative model does not know how to map the pixels from a static reference image to a completely new and moving posture in the generated video. To prove this, we introduce a per-frame render condition. This input supplies the appearance of the robot perfectly aligned to the target motion at every single time step. 

\begin{figure}[t]
\centering
\includegraphics[width=0.9\columnwidth]{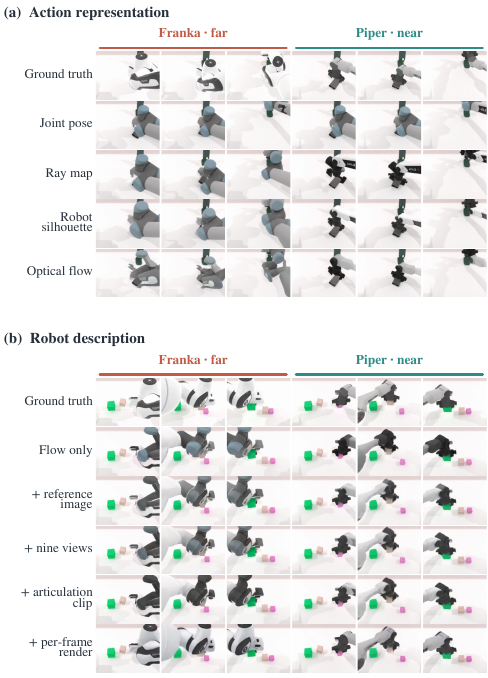}
\caption{\textbf{Pixel-space actions preserve structural coherence, while time-aligned robot cues recover identity.} 
(a) Pixel-aligned action representations retain the held-out robot more faithfully during movement than numeric joint actions. 
(b) On a distinct episode, static robot descriptions change appearance only modestly, whereas a per-frame render restores the target morphology.}
\label{fig:qual-actrep-cond}
\end{figure}

The empirical effect of time-aligned rendering is massive. Compared to using optical flow alone, the per-frame render cuts robot-region LPIPS by 50\% on Franka and 45\% on Piper. It also boosts shape IoU by 22\% and 8\% respectively. These improvements are several times larger than those achieved by providing nine static views. This confirms our hypothesis. Action-conditioned world models can successfully utilize robot appearance information only when that information is explicitly registered to the target frames over time. Figure~\ref{fig:qual-actrep-cond}(b) visualizes this difference. Static descriptions fail to change the generated output meaningfully, but the per-frame render guides the model to recover substantially more of the target morphology.

\begin{table}[t]
\centering
\scriptsize
\setlength{\tabcolsep}{2.0pt}
\renewcommand{\arraystretch}{1.12}
\begin{tabular}{@{}cl cccc c@{}}
\toprule
\multirow{2}{*}{\hd{Target}} &
\multirow{2}{*}{\hd{Metric}} &
\multicolumn{4}{c}{\hd{Demos per task} $M$} &
\multirow{2}{*}{\shortstack{\hd{Per-frame}\\\hd{render}}} \\
\cmidrule(lr){3-6}
& & $0$ & $1$ & $2$ & $3$ & \\
\midrule
\multirow{3}{*}{Franka}
& Sym.\ IoU\,$\uparrow$             & 0.679 & 0.748 & 0.757 & \best{0.758} & 0.815 \\
& Robot-region LPIPS\,$\downarrow$   & 0.230 & 0.158 & 0.142 & \best{0.136} & 0.126 \\
& Full-frame LPIPS\,$\downarrow$     & 0.167 & 0.117 & 0.107 & \best{0.102} & 0.106 \\
\addlinespace[2pt]
\multirow{3}{*}{Piper}
& Sym.\ IoU\,$\uparrow$             & 0.777 & 0.815 & 0.815 & \best{0.817} & 0.836 \\
& Robot-region LPIPS\,$\downarrow$   & 0.128 & 0.093 & 0.084 & \best{0.080} & 0.076 \\
& Full-frame LPIPS\,$\downarrow$     & 0.090 & 0.061 & 0.055 & \best{0.053} & 0.069 \\
\bottomrule
\end{tabular}
\caption{\textbf{Few-shot target adaptation.} $M=0$ is the zero-shot nine-view
checkpoint and the final column reports the per-frame render condition.}
\label{tab:fewshot}
\end{table}

\begin{figure}[t]
\centering
\captionsetup{skip=3pt}
\includegraphics[width=\columnwidth]{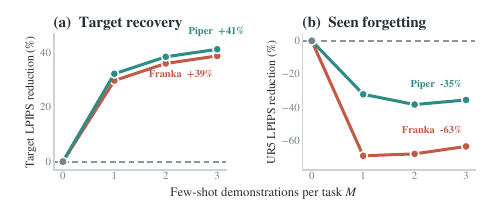}
\caption{\textbf{Adaptation benefit and forgetting.} Fine-tuning on the
held-out robot reduces its LPIPS error, but increases the LPIPS error on the
previously seen UR5 robot.}
\label{fig:fewshot-retention}
\end{figure}

\subsection{Can Few-Shot Adaptation Close the Gap Without Forgetting?}
\label{sec:exp-fewshot}

Given that zero-shot generalization using static embodiment's descriptions fails, we analyze whether training the model directly on a few target episodes can close the visual gap. We evaluate fine-tuning with $M \in \{1, 2, 3\}$ demonstrations per task. Table~\ref{tab:fewshot} demonstrates that fine-tuning produces rapid improvements. The very first round of demonstrations ($M=1$, representing just 25 episodes) delivers the vast majority of the benefit. It reduces LPIPS by 30\% on Franka and 32\% on Piper. Pushing the data to 75 episodes ($M=3$) only provides diminishing returns, bringing the total LPIPS reduction to roughly 40\%. 

However, relying entirely on full-frame metrics paints an overly optimistic picture. At $M=3$, Piper successfully closes 97\% of its global LPIPS gap. Yet, when we isolate the metric to just the robot region, it only closes 57\% of the gap. Franka shows a similar pattern, closing only 51\% of its region gap. This divergence indicates that the global scores are dominated by non-robot pixels such as lighting and background, while the network still struggles to render the precise mechanical details and textures of the new embodiment.

Furthermore, this adaptation introduces a severe functional cost. Figure~\ref{fig:fewshot-retention} tracks what happens to previously learned robots during this fine-tuning process. As the model adapts to Franka using just one demonstration per task, the LPIPS error on the previously seen UR5 robot spikes by 69\%. This catastrophic forgetting indicates that few-shot adaptation acts as a localized patch rather than a fundamental solution. The network overwrites its internal representation of the old robots to memorize the visual patterns of the new one. This trade-off suggests that deploying these models in diverse environments will require parameter-efficient tuning or replay mechanisms to preserve knowledge.

\begin{figure}[t]
\centering
\captionsetup{skip=3pt}
\includegraphics[width=\columnwidth]{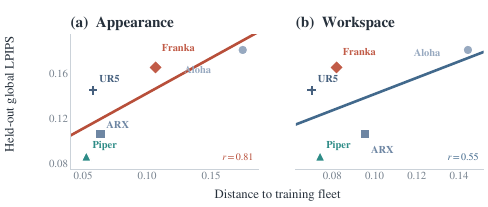}
\caption{\textbf{Correlation between embodiment distance and generalization error.}}
\label{fig:distance}
\end{figure}

\begin{table*}[t]
    \centering
    \footnotesize
    \setlength{\tabcolsep}{2.0pt}
    \renewcommand{\arraystretch}{1.25}
    \begin{tabular*}{0.8\textwidth}{@{\extracolsep{\fill}}ll cccc@{}}
    \toprule
    \multicolumn{1}{c}{\multirow{2}{*}{\hd{Model}}} &
    \multicolumn{1}{c}{\multirow{2}{*}{\hd{Action}}} &
    \hd{Visual} & \hd{Morphology} &
    \hd{Kinematics} & \hd{Object Dyn.} \\
    & & LPIPS\,$\downarrow$ & Sym.\ IoU\,$\uparrow$ &
    PCK\,$\uparrow$ & Traj.\ px.\,$\downarrow$ \\
    \midrule
    IRASim       & Rel.\ EE pose & 0.225\,/\,0.310\,/\,0.248 & 0.548\,/\,0.442\,/\,0.505 & 0.412\,/\,0.320\,/\,0.395 & 20.8\,/\,25.2\,/\,18.4 \\
    Ctrl-World   & Abs.\ EE pose & 0.168\,/\,0.262\,/\,0.185 & 0.654\,/\,0.528\,/\,0.612 & 0.468\,/\,0.392\,/\,0.452 & 16.5\,/\,18.6\,/\,14.8 \\
    EnerVerse-AC & EE $+$ ray maps  & 0.124\,/\,0.235\,/\,0.158 & 0.712\,/\,0.584\,/\,0.685 & 0.512\,/\,0.385\,/\,0.508 & 14.2\,/\,19.8\,/\,12.5 \\
    \midrule
    \textbf{\sysname{} ref.} & Optical flow & \textbf{0.056\,/\,0.180\,/\,0.097} & \textbf{0.817\,/\,0.667\,/\,0.773} & \textbf{0.596\,/\,0.456\,/\,0.593} & \textbf{9.9\,/\,16.1\,/\,8.7} \\
    \bottomrule
    \end{tabular*}
    \caption{\textbf{Cross-embodiment generalization gap across multiple world models.} Cells report the performance on seen fleet\,/\,Franka\,/\,Piper. The universal performance degradation on held-out robots confirms that this failure is a fundamental limitation across current action-conditioned world model architectures.}
    \label{tab:baseline}
    \end{table*}

\subsection{Does Embodiment Distance Track Generalization Difficulty?}
\label{sec:exp-distance}

Not all unseen robots pose the same challenge to the model. To empirically investigate whether the difference between the target robot and the training robots predicts this generation difficulty, we evaluate all five robots using a 5-fold leave-one-embodiment-out (LOEO) split. Under this protocol, we compare two distinct types of distances: visual appearance distance and physical reachable workspace distance. 

Figure~\ref{fig:distance} details the correlation between these distances and the resulting global LPIPS error. Across the five folds, we find a strong empirical correlation ($r=0.812$) between visual appearance distance and cross-embodiment generation error. Under a stability analysis, this positive trend remains consistent no matter which robot is held out. If a new embodiment looks drastically different from the training data, the model reliably struggles to render it.

In sharp contrast, the physical reachable workspace distance shows a weak and unstable relationship ($r=0.549$), failing the stability analysis across different folds. This empirical contrast exposes a core characteristic of current video generation backbones. The network behaves primarily as a 2D visual pattern matcher, not as a system that genuinely understands physical kinematics. Therefore, a robot with a radically different mechanical structure is not harder for the model to generate unless it also lacks visual familiarity.

\subsection{Does the Generalization Gap Persist Across World Models?}
\label{sec:exp-baseline}

To establish that the observed generalization failure is not an isolated artifact of our reference model, we evaluate three other recent world models on our dataset: IRASim \citep{irasim}, Ctrl-World \citep{ctrlworld}, and EnerVerse-AC \citep{enerverse}, all of which are fine-tuned on our testbed data from their respective pre-trained checkpoints. Table~\ref{tab:baseline} confirms that the visual degradation on unseen embodiments is a shared limitation spanning diverse state-of-the-art architectures.

Every single model experiences a noticeable performance drop when required to generate the unseen Franka and Piper robots. For example, when attempting to generate Franka, the LPIPS error increases between 38\% and 221\% relative to each model's baseline seen performance. Simultaneously, their shape IoU metrics drop consistently by nearly 20\%. 

This universal degradation proves that the cross-embodiment visual gap stems from the fundamental way current video generation pipelines process action and structure. Notably, this cross-model comparison corroborates our earlier findings: models incorporating pixel-space action conditioning (such as optical flow in our reference model or ray maps in EnerVerse-AC) achieve stronger absolute performance on held-out robots than those relying purely on abstract numeric poses (IRASim and Ctrl-World). While architectures with stronger baseline capabilities generate cleaner images overall, they still suffer proportional degradation when encountering a new embodiment. This shared bottleneck confirms that merely scaling existing 2D generative priors is insufficient. Achieving true cross-embodiment simulation demands new architectures that explicitly decouple visual appearance from underlying physical kinematics.

\subsection{What Are the Design Prerequisites for Future World Models?}
\label{sec:discussion}

Our empirical study clearly shows that current action-conditioned world models lack robust generalization to new embodiments. The analysis isolates the core reasons behind this failure. Models do not just need static visual descriptions of the target robot. They need inputs that establish explicit temporal and spatial correspondence across frames. Without time-aligned spatial inputs, the generative backbone simply guesses the physical articulation based on 2D visual patterns.

These findings suggest several potential directions for future research. First, cross-embodiment models may benefit from transitioning action representations from abstract numeric joint actions to pixel-space representations, such as optical flow or projected ray maps, which align naturally with the visual space. Second, solving the alignment barrier between static embodiment's descriptions and dynamic motion is crucial. Future models might explore mechanisms that persistently bind explicit geometric or structural conditioning to the target morphology throughout the video rollout. Finally, overcoming this fundamental generalization bottleneck requires architectural innovations that explicitly decouple high-level visual appearance generation from underlying physical dynamics and kinematic constraints.
\section{Conclusion}

We introduced \sysname{} to investigate whether action-conditioned world models genuinely learn physical kinematics or merely memorize visual patterns. By isolating the robot embodiment in controlled evaluations, we demonstrate that current architectures behave predominantly as 2D visual pattern matchers. Their cross-embodiment generalization is fundamentally bottlenecked by visual similarity rather than physical kinematic similarity. This intrinsic limitation explains their struggle to predict dynamic visual changes from static initial observations conditioned on abstract actions, as well as the catastrophic forgetting triggered by few-shot adaptation. Taken together, these findings highlight that achieving true cross-embodiment generalization requires moving beyond 2D priors toward architectural innovations that decouple visual appearance from underlying physical dynamics.

{\small
\bibliography{robotwin-xe}}

\clearpage
\appendix
\raggedbottom
\section{Scope and Reporting Convention}

This appendix reports the complete metric suite for the
controlled distance, action-representation, conditioning, and few-shot studies.
Results from the fixed three-train / two-held-out protocol and the five-fold
leave-one-embodiment-out (LOEO) protocol are kept separate because the
corresponding models are trained on different robot fleets.  Unless stated otherwise, values are episode means.  We
follow the four evaluation dimensions defined in the main paper and report them
side by side rather than collapsing them into a single scalar: visual quality
(LPIPS, PSNR, and SSIM), robot morphology (symmetric IoU, boundary $F_1$, and
region LPIPS), robot kinematics (PCK, keypoint error, and nDTW against
URDF-derived forward kinematics), and object dynamics (object-trajectory error
against the simulator's recorded object poses).  As stated in the main paper,
the kinematics and object-dynamics dimensions are scored in the image plane
using CoTracker3 correspondences, so they measure projected motion rather than
recovered joint angles.  PCK denotes PCK at $\alpha=0.1$.

\FloatBarrier

\section{Dataset Composition and Evaluation Protocols}

\subsection{Paired dataset}

The dataset contains 25 manipulation tasks and five bimanual robot setups:
Aloha-AgileX, Arx-X5, UR5, Franka Panda, and Piper.  For a fixed task and seed,
scene geometry, object initial states, lighting, and camera parameters are held
constant across embodiments.  Each episode stores RGB frames, joint actions,
camera intrinsics and extrinsics, rigid and articulated object poses, contact
states, and a robot-only render used to obtain the ground-truth mask.  Native
frames are $320\times240$ pixels.

\begin{table}[t]
\centering
\footnotesize
\setlength{\tabcolsep}{3.2pt}
\renewcommand{\arraystretch}{1.12}
\begin{tabular}{@{}lcccc@{}}
\toprule
\hd{Split} & \hd{Tasks} & \hd{Robots} & \hd{Seeds/task} & \hd{Candidate eps.} \\
\midrule
Training core & 25 & 5 & 50 & 6,250 \\
Disjoint test & 25 & 5 & 20 & 2,500 \\
\bottomrule
\end{tabular}
\caption{\textbf{Dataset composition.} Candidate episodes count every
task-seed-robot cell before the success/validity filter.  The committed
fixed-split records contain 2,500 scored cells: 500 for each of the five
robots.}
\label{tab:data-composition}
\end{table}

\begin{table*}[t]
\centering
\footnotesize
\setlength{\tabcolsep}{3.4pt}
\renewcommand{\arraystretch}{1.12}
\begin{tabular}{@{}lllll@{}}
\toprule
\multicolumn{5}{c}{\hd{Manipulation tasks}} \\
\midrule
Beat block with hammer & Blocks ranking (RGB) & Blocks ranking (size) & Grab roller & Lift pot \\
Move can to pot & Move pill bottle to pad & Move stapler to pad & Open laptop & Place A-to-B (left) \\
Place A-to-B (right) & Place bread in basket & Place burger with fries & Place container on plate & Place dual shoes \\
Place empty cup & Place mouse on pad & Place object on scale & Place object on stand & Place phone on stand \\
Press stapler & Shake bottle & Shake bottle horizontally & Stack two bowls & Stamp seal \\
\bottomrule
\end{tabular}
\caption{\textbf{The 25-task evaluation suite.} Every protocol uses the same
task set. Training/adaptation and evaluation seeds are disjoint.}
\label{tab:task-list}
\end{table*}

\subsection{Fixed split, LOEO, and few-shot protocols}

The fixed split trains on Aloha-AgileX, Arx-X5, and UR5, then evaluates Franka
and Piper separately.  It scores 1,500 seen-robot episodes, 500 Franka
episodes, and 500 Piper episodes.  The action-representation and
conditioning experiments use this protocol.

The distance study uses five LOEO folds.  In each fold, one robot is held out
and the other four are used for training.  Every fold uses the same 2,500
valid task-seed-robot cells.  In every fold, the split is exactly 2,000 seen
and 500 held-out records.
Absolute fixed-split and LOEO scores should therefore not be compared directly.

Few-shot models resume from the fixed-split multi-view checkpoint.  For a
target robot, $M\in\{1,2,3\}$ demonstrations are sampled \emph{per task}, for
25, 50, or 75 adaptation episodes in total.  Target episodes measure recovery, while
500 disjoint UR5 episodes measure forgetting.

\begin{table}[t]
\centering
\footnotesize
\setlength{\tabcolsep}{2.8pt}
\renewcommand{\arraystretch}{1.12}
\begin{tabular}{@{}lll@{}}
\toprule
\hd{Experiment} & \hd{Seen / reference} & \hd{Held-out / target} \\
\midrule
Action representation & 1,500 pooled & Franka 500, Piper 500 \\
Conditioning & 1,500 pooled & Franka 500, Piper 500 \\
LOEO fold & 2,000 pooled & 500 held-out \\
Few-shot & UR5 500 & Franka 500, Piper 500 \\
\bottomrule
\end{tabular}
\caption{\textbf{Numbers of scored episodes.} Conditions within each
controlled comparison are paired on their shared task-seed-robot cells.}
\label{tab:eval-counts}
\end{table}

\section{Metric Definitions and Evidence Hierarchy}

\subsection{Visual Quality}

PSNR and SSIM are computed per frame against the ground-truth RGB rollout and
then averaged over time.  PSNR operates on RGB values in $[0,1]$. Exact frames
are capped at the largest finite per-frame PSNR when mixed with non-exact
frames.  SSIM uses the standard channel-aware structural-similarity
implementation.  LPIPS uses the AlexNet-backed LPIPS network
\citep{lpips}, averages over frames, and is lower-is-better.

\subsection{Robot Morphology}

The ground-truth mask comes from the byte-aligned robot-only render.  SAM2
\citep{sam2} is initialized with that mask at the frame containing the largest
visible robot region and propagates a mask bidirectionally over the predicted
video.  Symmetric IoU penalizes both missed robot pixels and hallucinated robot
pixels.  Boundary $F_1$ matches predicted and ground-truth contour pixels with
a 2-pixel tolerance.  Region LPIPS is evaluated inside the bounding box of the
union of valid ground-truth robot masks.

\subsection{Robot Kinematics}

Ground-truth keypoints are link positions obtained by replaying each frame's
joint action through the robot URDF, applying forward kinematics, and projecting
the link positions into the head camera.  At the frame with the most visible FK
keypoints, these points initialize CoTracker3 \citep{cotracker3} on the
\emph{predicted} video.  Metrics use only FK-visible and tracker-confident
point-frame pairs.

Let $d_t$ be the diagonal of the bounding box of all visible FK keypoints at
frame $t$.  The reported PCK is
\begin{equation}
\mathrm{PCK}_{0.1}=\frac{1}{N}\sum_{t,i}
\mathbb{1}\!\left[\lVert\hat{p}_{t,i}-p_{t,i}\rVert_2
<0.1d_t\right].
\end{equation}
The evaluator also computes PCK at $\alpha=0.05$. Complete tables consistently
show $\alpha=0.1$.  Keypoint pixel error is the mean Euclidean reprojection
error.  For each keypoint, nDTW applies classic dynamic time warping to its
predicted and FK trajectories and divides the accumulated cost by the sum of
the two sequence lengths and their mean visible-keypoint bounding-box diagonal.

\subsection{Object Dynamics}

CoTracker3 tracks object locations in the predicted rollout, while simulator
object poses projected into the head camera provide the ground-truth image
trajectory.  Object-trajectory pixel error is the mean Euclidean pixel-space
error over frames where both tracks are valid and the object is visible, then
averaged across tracked objects.  Contact $F_1$ and related diagnostics are
computed internally but excluded from the reported suite because calibration
tests found contact timing insufficiently reliable.

\subsection{Embodiment distances}

Kinematic-workspace distance is training-free.  For each embodiment, 1,500
joint configurations are sampled uniformly within the arm limits.  Forward
kinematics yields the origins of every link in the arm-base frame. At most 4,000
points are retained.  Pairwise distance is the symmetric Chamfer distance
between the two workspace clouds.

Appearance distance uses the nine standardized robot-only renders.  The run
reported here used the recorded \emph{HSV-histogram plus silhouette Hu-moment}
backend: a normalized $16\times16$ hue-saturation histogram is concatenated
with seven log-transformed Hu moments for each view, descriptors are averaged
over views, and pairwise cosine distance is computed.  Each LOEO robot's scalar
distance is its mean pairwise distance to the other four robots.

\FloatBarrier

\section{Model Architecture and Training Details}
\label{sec:supp-training}

\subsection{Backbone and conditioning streams}

Our reference model adapts the FlowWAM design \citep{flowwam} using the
Wan2.2-TI2V-5B video diffusion backbone \citep{wan}.  The pretrained VAE maps
all image-like inputs to video latents, and the UMT5 encoder supplies the task
text context.  Both encoders remain frozen.  We fine-tune the complete DiT
rather than using LoRA, together with the trainable input modules introduced
for the action and embodiment streams.

The model predicts the RGB rollout from a clean first-frame latent.  Future RGB
latents are noised according to the flow-matching schedule, whereas action and
embodiment inputs remain clean conditioning.  Pixel-space action inputs use a
patch embedding copied from the pretrained RGB patch embedding and a learned
stream-type embedding.  The optional embodiment stream uses a separate copy of
the same patch embedding but has no decoder head.  For the numeric joint-pose
ablation, a two-layer MLP with a SiLU nonlinearity maps the canonical
16-dimensional action vector to one DiT-width token per latent frame.

After patchification, RGB, action, and optional embodiment tokens are assigned
independent spatiotemporal rotary positions and concatenated for joint
self-attention.  Each stream then passes through the same text cross-attention
and feed-forward weights in every DiT block.  The final embodiment tokens (or
numeric action tokens) are discarded: these streams influence RGB prediction
only through attention.  This construction keeps the transformer backbone
fixed across the visual action-representation and conditioning ablations.

\begin{table}[t]
\centering
\scriptsize
\setlength{\tabcolsep}{2.8pt}
\renewcommand{\arraystretch}{1.14}
\begin{tabular}{@{}p{0.19\columnwidth}p{0.47\columnwidth}p{0.27\columnwidth}@{}}
\toprule
\hd{Stream} & \hd{Input and encoder} & \hd{Role} \\
\midrule
RGB & video latent. Pretrained DiT patch embedding & noised prediction target \\
Image action & RAFT flow, mask, or ray map. Copied patch embedding & clean conditioning \\
Embodiment & image, views, clip, or per-frame render. Copied patch embedding & clean conditioning \\
Numeric action & canonical 16-D joints. Linear-SiLU-Linear & clean conditioning \\
\bottomrule
\end{tabular}
\caption{\textbf{Reference-model streams.} All streams share the DiT blocks.
only the RGB rollout is directly supervised.}
\label{tab:stream-structure}
\end{table}

\subsection{Action and embodiment inputs}

Robot-masked optical flow is computed with RAFT \citep{raft}, clipped to a
maximum magnitude of 25, and cached before training.  It is computed on the
robot-only forward-kinematics render of the commanded action sequence, not on
the ground-truth scene video.  Given the action sequence, the target URDF, and
the camera parameters, this stream is therefore derivable at inference time
without observing the future rollout.  The mask stream is the per-frame binary
silhouette of that same render.  The ray
map projects the end-effector frame into the head camera, drawing the three
coordinate axes and a base marker whose color records the gripper state.  The
numeric representation zero-pads each arm to seven joints and appends two
gripper values, yielding the shared 16-D layout for both 6- and 7-DoF
embodiments.

The task-independent embodiment conditions contain either one canonical
front-view image, nine uniformly rendered azimuth views, or a 17-frame
articulation clip.  The per-frame rendering condition instead supplies a
full-length robot render produced by forward kinematics.  Short
specifications are reused in every autoregressive chunk. The per-frame render is
sliced to match the corresponding rollout frames.  All specification inputs
are encoded once into clean VAE latents and never receive a reconstruction
objective.

\subsection{Optimization and rollout training}

All fixed-split and LOEO runs start from the same Wan2.2-TI2V-5B weights and use
50 core episodes per task and robot.  We
optimize the DiT and the active stream modules with AdamW, weight decay 0.01,
and a constant learning rate of $5\times10^{-5}$.  Training uses bfloat16,
gradient checkpointing, one sample per GPU, and one gradient-accumulation step
on a single eight-GPU node.  Thus the distributed batch contains eight episode
samples, although a sample may contain one or two autoregressive chunks.

Each input contains 121 frames at $320\times256$.  From native $320\times240$
frames, the loader emits the latter resolution so both dimensions are divisible
by the VAE-DiT spatial factor. For a second
chunk, the last decoded
RGB prediction from the preceding chunk becomes the next clean reference
frame.  The reference latent receives Gaussian perturbation with a randomly
sampled strength between 0 and 0.1 during training.  VAE latents and text
embeddings are cached per condition. The predicted reference for later chunks
is necessarily encoded online.

Although the pixel-space action module inherits a copied prediction head,
all reported experiments set its loss weight to zero.  The optimized objective
is therefore the mean flow-matching MSE on future RGB latents, averaged equally
over the sampled rollout chunks.  This prevents an auxiliary flow-reconstruction
objective from changing across action representations.

\begin{table}[t]
\centering
\scriptsize
\setlength{\tabcolsep}{3.0pt}
\renewcommand{\arraystretch}{1.14}
\begin{tabular}{@{}p{0.25\columnwidth}p{0.38\columnwidth}cc@{}}
\toprule
\hd{Stage} & \hd{Initialization} & \hd{Epochs} & \hd{LR} \\
\midrule
Fixed split & Wan2.2-TI2V-5B & 40 & $5\!\times\!10^{-5}$ \\
LOEO fold & Wan2.2-TI2V-5B & 40 & $5\!\times\!10^{-5}$ \\
Few-shot & multi-view, epoch 39 & 20 & $2\!\times\!10^{-5}$ \\
\bottomrule
\end{tabular}
\caption{\textbf{Training schedules.} Epoch indices start at zero, so the
40-epoch checkpoint is epoch 39.}
\label{tab:training-schedules}
\end{table}

\subsection{Few-shot adaptation}

Few-shot runs initialize every DiT, action-stream, and embodiment-stream weight
from the fixed-split multi-view checkpoint at epoch 39.  For each target robot,
$M\in\{1,2,3\}$ training episodes are selected independently for each of the 25
tasks, giving 25, 50, or 75 adaptation episodes.  These demonstrations use the
training seed family and remain disjoint from all reported test episodes.  We
hold the budget at 20 passes over the adaptation set for every $M$, reduce the
learning rate to $2\times10^{-5}$, and otherwise retain the optimizer, precision,
batching, rollout, and RGB-only loss settings above.  Adaptation uses only the
target robot and no replay from the original three-robot training fleet. The
separate UR5 evaluation therefore measures the resulting forgetting directly.

\FloatBarrier

\section{Complete Embodiment-Distance Results}

Tables~\ref{tab:robot-distance-features}-\ref{tab:pairwise-distances} report
the per-robot distance features and both complete distance matrices.  The remaining
tables give every held-out score and the corresponding correlation analysis.
For every metric, signs are oriented so that positive means ``farther implies
worse.''  The sensitivity interval is the minimum and maximum Pearson
correlation obtained after removing each robot in turn. An interval crossing
zero indicates sign instability.

Global LPIPS is the primary distance outcome.  Appearance distance gives
$r=0.812$, exact two-sided permutation $p=0.075$, and leave-one-robot-out range
$[0.760,0.923]$.  Kinematic workspace distance gives $r=0.549$, $p=0.392$, and
range $[-0.155,0.706]$.  All other rows are exploratory.  The PCK, keypoint,
nDTW, and object-trajectory rows follow the kinematics and object-dynamics
dimensions of the main paper and are therefore scored in the image plane, as
described in the protocol.

\begin{table*}[t]
\centering
\scriptsize
\setlength{\tabcolsep}{2.0pt}
\renewcommand{\arraystretch}{1.12}
\begin{tabular}{@{}lccccc@{}}
\toprule
\hd{Robot} & \hd{DoF} & \hd{Links} & \shortstack{\hd{Reach}\\\hd{(m)}} &
\shortstack{\hd{Volume}\\\hd{(m$^3$)}} & \shortstack{\hd{Mean LOEO distance}\\\hd{kin. / app.}} \\
\midrule
ALOHA-AgileX & 6 & 54 & 1.639 & 1.374 & 0.144 / 0.175 \\
ARX-X5       & 6 & 11 & 0.803 & 1.133 & 0.095 / 0.063 \\
UR5          & 6 & 18 & 1.083 & 2.771 & 0.070 / 0.058 \\
Franka       & 7 & 14 & 1.262 & 2.607 & 0.082 / 0.107 \\
Piper        & 6 & 11 & 0.885 & 1.326 & 0.074 / 0.052 \\
\bottomrule
\end{tabular}
\caption{\textbf{Kinematic features and LOEO distances.} Reach and convex-hull
workspace volume are diagnostics. Only the mean workspace-Chamfer and
appearance-cosine distances to the other four robots are used as correlation
axes.}
\label{tab:robot-distance-features}
\end{table*}

\begin{table*}[t]
\centering
\footnotesize
\begin{minipage}[t]{0.485\textwidth}
\centering
\setlength{\tabcolsep}{4.2pt}
\renewcommand{\arraystretch}{1.08}
\begin{tabular}{@{}lccccc@{}}
\toprule
\multicolumn{6}{c}{\hd{Kinematic workspace Chamfer distance (m)}} \\
\midrule
 & \hd{Aloha} & \hd{Arx} & \hd{UR5} & \hd{Franka} & \hd{Piper} \\
Aloha  & 0     & 0.195 & 0.123 & 0.111 & 0.147 \\
Arx    & 0.195  & 0    & 0.059 & 0.094 & 0.034 \\
UR5    & 0.123  & 0.059 & 0    & 0.053 & 0.046 \\
Franka & 0.111  & 0.094 & 0.053 & 0    & 0.070 \\
Piper  & 0.147  & 0.034 & 0.046 & 0.070 & 0 \\
\bottomrule
\end{tabular}
\end{minipage}
\hfill
\begin{minipage}[t]{0.485\textwidth}
\centering
\setlength{\tabcolsep}{4.2pt}
\renewcommand{\arraystretch}{1.08}
\begin{tabular}{@{}lccccc@{}}
\toprule
\multicolumn{6}{c}{\hd{Appearance-descriptor cosine distance}} \\
\midrule
 & \hd{Aloha} & \hd{Arx} & \hd{UR5} & \hd{Franka} & \hd{Piper} \\
Aloha  & 0     & 0.186 & 0.137 & 0.235 & 0.139 \\
Arx    & 0.186  & 0    & 0.012 & 0.050 & 0.005 \\
UR5    & 0.137  & 0.012 & 0    & 0.078 & 0.003 \\
Franka & 0.235  & 0.050 & 0.078 & 0    & 0.062 \\
Piper  & 0.139  & 0.005 & 0.003 & 0.062 & 0 \\
\bottomrule
\end{tabular}
\end{minipage}
\caption{\textbf{Complete pairwise embodiment-distance matrices.} The matrices
are symmetric.  LOEO axis values in Table~\ref{tab:robot-distance-features}
are row means excluding the diagonal.}
\label{tab:pairwise-distances}
\end{table*}

\begin{table*}[t]
\centering
\scriptsize
\setlength{\tabcolsep}{1.35pt}
\renewcommand{\arraystretch}{1.16}
\begin{tabular*}{\textwidth}{@{\extracolsep{\fill}}l ccc ccc ccc c@{}}
\toprule
\multicolumn{1}{c}{\multirow{2}{*}[-0.5ex]{\hd{Held-out robot}}} & \multicolumn{3}{c}{\hd{Visual Quality}}
& \multicolumn{3}{c}{\hd{Robot Morphology}}
& \multicolumn{3}{c}{\hd{Robot Kinematics}}
& \hd{Object Dynamics} \\
\cmidrule(lr){2-4}\cmidrule(lr){5-7}\cmidrule(lr){8-10}\cmidrule(l){11-11}
& PSNR\,$\uparrow$ & SSIM\,$\uparrow$ & LPIPS\,$\downarrow$
& Sym.\ IoU\,$\uparrow$ & Boundary $F_1$\,$\uparrow$ & Region LPIPS\,$\downarrow$
& PCK\,$\uparrow$ & Kpt.\ px.\ error\,$\downarrow$ & nDTW\,$\downarrow$
& Trajectory px.\ error\,$\downarrow$ \\
\midrule
ALOHA-AgileX & 18.791 & 0.855 & 0.181 & 0.854 & 0.796 & 0.097 & 0.443 &  7.424 & 0.077 & 15.214 \\
ARX-X5       & 21.306 & 0.854 & 0.107 & 0.778 & 0.676 & 0.124 & 0.834 &  9.851 & 0.029 &  8.852 \\
Franka       & 19.429 & 0.850 & 0.166 & 0.659 & 0.471 & 0.231 & 0.463 & 21.914 & 0.109 & 15.847 \\
Piper        & 25.910 & 0.912 & 0.087 & 0.785 & 0.684 & 0.124 & 0.592 & 12.999 & 0.088 &  8.382 \\
UR5          & 22.143 & 0.863 & 0.145 & 0.824 & 0.703 & 0.163 & 0.578 & 15.718 & 0.066 & 14.797 \\
\bottomrule
\end{tabular*}
\caption{\textbf{Complete held-out absolute scores for the five LOEO folds.}
Each row comes from a separately trained four-robot model.  PCK uses
$\alpha=0.1$.}
\label{tab:loeo-absolute}
\end{table*}

\begin{table*}[t]
\centering
\small
\setlength{\tabcolsep}{5.0pt}
\renewcommand{\arraystretch}{1.12}
\begin{tabular}{@{}lccccc@{}}
\toprule
\multirow{2}{*}{\hd{Metric}} & \multicolumn{2}{c}{\hd{Appearance distance}} &
\multicolumn{2}{c}{\hd{Kinematic distance}} & \multirow{2}{*}{\hd{DoF $r$}} \\
\cmidrule(lr){2-3}\cmidrule(lr){4-5}
& \hd{Pearson $r$} & \hd{LOO range} & \hd{Pearson $r$} & \hd{LOO range} & \\
\midrule
PSNR                & +0.769 & [+0.780,+0.918] & +0.633 & [+0.443,+0.809] & +0.416 \\
SSIM                & +0.473 & [+0.393,+0.604] & +0.390 & [+0.265,+0.509] & +0.360 \\
LPIPS               & +0.812 & [+0.760,+0.923] & +0.549 & [$-$0.155,+0.706] & +0.402 \\
Symmetric IoU       & $-$0.176 & [$-$0.813,+0.952] & $-$0.448 & [$-$0.691,+0.295] & +0.911 \\
Boundary $F_1$      & $-$0.231 & [$-$0.972,+0.984] & $-$0.534 & [$-$0.872,+0.177] & +0.915 \\
Region LPIPS        & $-$0.175 & [$-$0.736,+0.917] & $-$0.552 & [$-$0.833,$-$0.135] & +0.893 \\
PCK ($\alpha=.1$)   & +0.645 & [+0.533,+0.918] & +0.282 & [$-$0.685,+0.748] & +0.428 \\
Keypoint px. error  & $-$0.286 & [$-$0.775,+0.823] & $-$0.671 & [$-$0.913,$-$0.358] & +0.829 \\
nDTW                & +0.286 & [+0.065,+0.572] & $-$0.124 & [$-$0.546,+0.022] & +0.671 \\
Trajectory px. error& +0.611 & [+0.492,+0.830] & +0.267 & [$-$0.364,+0.479] & +0.491 \\
\bottomrule
\end{tabular}
\caption{\textbf{Complete exploratory distance analysis.} LOO ranges expose
the sensitivity of five-point correlations.  The DoF column is diagnostic
only: Franka is the sole 7-DoF robot.}
\label{tab:distance-correlations}
\end{table*}

\FloatBarrier

\onecolumn
\section{Complete Intervention Results}

\noindent\begin{minipage}[t]{0.485\textwidth}
This section reports every metric for the action-representation, conditioning,
few-shot-recovery, and forgetting experiments, including the PSNR, SSIM,
kinematics, and object-dynamics columns that the main paper omits for space.

The action interfaces differ only in the second conditioning stream: joint
pose is a zero-padded 16-dimensional bimanual joint vector. The ray map projects
the end-effector frame and gripper state into the image. The mask supplies a
per-frame binary robot map, and optical flow supplies robot-masked RAFT flow.
The mask and flow streams are both computed from the robot-only
forward-kinematics render of the commanded action sequence, so neither observes
the ground-truth scene video.  In the
conditioning study, the per-frame rendering condition provides the target
robot's forward-kinematics render at each time step.
\end{minipage}\hfill
\begin{minipage}[t]{0.485\textwidth}

Few-shot models resume from the multi-view checkpoint, and forgetting is
measured on the same 500 disjoint UR5 episodes after every target adaptation.
Recovery and forgetting are reported in separate tables because they evaluate
different robot sets and answer different questions.
\end{minipage}

\begin{table}[H]
\centering
\scriptsize
\setlength{\tabcolsep}{1.35pt}
\renewcommand{\arraystretch}{1.10}
\begin{tabular*}{\textwidth}{@{\extracolsep{\fill}}l ccc ccc ccc c@{}}
\toprule
\multicolumn{1}{c}{\multirow{2}{*}[-0.5ex]{\hd{Representation}}} & \multicolumn{3}{c}{\hd{Visual Quality}}
& \multicolumn{3}{c}{\hd{Robot Morphology}}
& \multicolumn{3}{c}{\hd{Robot Kinematics}}
& \hd{Object Dynamics} \\
\cmidrule(lr){2-4}\cmidrule(lr){5-7}\cmidrule(lr){8-10}\cmidrule(l){11-11}
& PSNR\,$\uparrow$ & SSIM\,$\uparrow$ & LPIPS\,$\downarrow$
& Sym.\ IoU\,$\uparrow$ & Boundary $F_1$\,$\uparrow$ & Region LPIPS\,$\downarrow$
& PCK\,$\uparrow$ & Kpt.\ px.\ error\,$\downarrow$ & nDTW\,$\downarrow$
& Trajectory px.\ error\,$\downarrow$ \\
\midrule
\multicolumn{11}{c}{\hd{Seen pool: Aloha-AgileX, Arx-X5, and UR5}} \\
Joint pose   & 22.035 & 0.870 & 0.101 & 0.709 & 0.554 & 0.114 & 0.520 & 15.051 & 0.078 & 10.551 \\
Ray map      & 25.598 & 0.907 & 0.055 & 0.789 & 0.661 & 0.061 & 0.596 & 11.209 & 0.062 &  9.147 \\
Mask         & 27.134 & 0.918 & 0.052 & 0.822 & 0.729 & 0.044 & 0.604 & 11.065 & 0.062 &  9.833 \\
Optical flow & 27.171 & 0.919 & 0.056 & 0.817 & 0.716 & 0.046 & 0.596 & 11.504 & 0.062 &  9.887 \\
\addlinespace[2pt]
\multicolumn{11}{c}{\hd{Held-out Franka}} \\
Joint pose   & 17.364 & 0.813 & 0.254 & 0.534 & 0.306 & 0.356 & 0.280 & 32.377 & 0.163 & 15.395 \\
Ray map      & 18.202 & 0.825 & 0.221 & 0.634 & 0.379 & 0.316 & 0.456 & 17.737 & 0.095 & 14.299 \\
Mask         & 18.376 & 0.836 & 0.207 & 0.818 & 0.636 & 0.299 & 0.492 & 16.709 & 0.090 & 16.190 \\
Optical flow & 19.191 & 0.843 & 0.180 & 0.667 & 0.463 & 0.251 & 0.456 & 21.557 & 0.111 & 16.067 \\
\addlinespace[2pt]
\multicolumn{11}{c}{\hd{Held-out Piper}} \\
Joint pose   & 18.554 & 0.837 & 0.228 & 0.358 & 0.208 & 0.353 & 0.219 & 39.099 & 0.245 & 15.376 \\
Ray map      & 21.013 & 0.863 & 0.167 & 0.530 & 0.328 & 0.271 & 0.584 & 12.512 & 0.080 & 10.834 \\
Mask         & 24.463 & 0.898 & 0.105 & 0.807 & 0.721 & 0.148 & 0.633 & 11.974 & 0.076 &  8.201 \\
Optical flow & 24.245 & 0.904 & 0.097 & 0.773 & 0.659 & 0.138 & 0.593 & 12.517 & 0.080 &  8.684 \\
\bottomrule
\end{tabular*}
\caption{\textbf{Complete action-representation evaluation.} The seen and two
held-out splits are reported separately for all ten metrics spanning the four
evaluation dimensions.}
\label{tab:actrep-full}
\end{table}

\begin{table}[H]
\centering
\scriptsize
\setlength{\tabcolsep}{1.35pt}
\renewcommand{\arraystretch}{1.08}
\begin{tabular*}{\textwidth}{@{\extracolsep{\fill}}l ccc ccc ccc c@{}}
\toprule
\multicolumn{1}{c}{\multirow{2}{*}[-0.5ex]{\hd{Conditioning}}} & \multicolumn{3}{c}{\hd{Visual Quality}}
& \multicolumn{3}{c}{\hd{Robot Morphology}}
& \multicolumn{3}{c}{\hd{Robot Kinematics}}
& \hd{Object Dynamics} \\
\cmidrule(lr){2-4}\cmidrule(lr){5-7}\cmidrule(lr){8-10}\cmidrule(l){11-11}
& PSNR\,$\uparrow$ & SSIM\,$\uparrow$ & LPIPS\,$\downarrow$
& Sym.\ IoU\,$\uparrow$ & Boundary $F_1$\,$\uparrow$ & Region LPIPS\,$\downarrow$
& PCK\,$\uparrow$ & Kpt.\ px.\ error\,$\downarrow$ & nDTW\,$\downarrow$
& Trajectory px.\ error\,$\downarrow$ \\
\midrule
\multicolumn{11}{c}{\hd{Seen pool: Aloha-AgileX, Arx-X5, and UR5}} \\
flow only        & 27.171 & 0.919 & 0.056 & 0.817 & 0.716 & 0.046 & 0.596 & 11.504 & 0.062 & 9.887 \\
$+$ ref. image   & 27.141 & 0.919 & 0.053 & 0.816 & 0.718 & 0.045 & 0.598 & 11.301 & 0.063 & 9.916 \\
$+$ multi-view   & 27.215 & 0.919 & 0.052 & 0.817 & 0.718 & 0.044 & 0.600 & 11.154 & 0.062 & 9.803 \\
$+$ artic. clip  & 27.165 & 0.918 & 0.052 & 0.817 & 0.717 & 0.044 & 0.602 & 11.159 & 0.061 & 9.784 \\
per-frame render & 27.281 & 0.918 & 0.051 & 0.826 & 0.736 & 0.043 & 0.608 & 11.128 & 0.064 & 9.828 \\
\addlinespace[2pt]
\multicolumn{11}{c}{\hd{Held-out Franka}} \\
flow only        & 19.191 & 0.843 & 0.180 & 0.667 & 0.463 & 0.251 & 0.456 & 21.557 & 0.111 & 16.067 \\
$+$ ref. image   & 19.788 & 0.849 & 0.168 & 0.667 & 0.468 & 0.231 & 0.458 & 21.586 & 0.112 & 16.320 \\
$+$ multi-view   & 19.661 & 0.850 & 0.167 & 0.679 & 0.476 & 0.230 & 0.463 & 21.347 & 0.109 & 15.831 \\
$+$ artic. clip  & 18.781 & 0.845 & 0.170 & 0.671 & 0.473 & 0.235 & 0.469 & 21.929 & 0.111 & 16.360 \\
per-frame render & 23.046 & 0.879 & 0.106 & 0.815 & 0.678 & 0.126 & 0.494 & 17.374 & 0.088 & 16.209 \\
\addlinespace[2pt]
\multicolumn{11}{c}{\hd{Held-out Piper}} \\
flow only        & 24.245 & 0.904 & 0.097 & 0.773 & 0.659 & 0.138 & 0.593 & 12.517 & 0.080 & 8.684 \\
$+$ ref. image   & 25.827 & 0.909 & 0.090 & 0.775 & 0.662 & 0.128 & 0.584 & 13.459 & 0.085 & 8.129 \\
$+$ multi-view   & 25.778 & 0.908 & 0.090 & 0.777 & 0.662 & 0.128 & 0.592 & 12.482 & 0.083 & 8.285 \\
$+$ artic. clip  & 25.431 & 0.906 & 0.092 & 0.772 & 0.656 & 0.130 & 0.580 & 13.364 & 0.088 & 8.278 \\
per-frame render & 27.995 & 0.922 & 0.069 & 0.836 & 0.776 & 0.076 & 0.615 & 12.101 & 0.070 & 7.638 \\
\bottomrule
\end{tabular*}
\caption{\textbf{Complete conditioning evaluation.} All conditions use the
same fixed split and evaluation cells.}
\label{tab:cond-full}
\end{table}

\begin{table}[H]
\centering
\scriptsize
\setlength{\tabcolsep}{1.35pt}
\renewcommand{\arraystretch}{1.10}
\begin{tabular*}{\textwidth}{@{\extracolsep{\fill}}l ccc ccc ccc c@{}}
\toprule
\multicolumn{1}{c}{\multirow{2}{*}[-0.5ex]{\hd{Demos per task}}} & \multicolumn{3}{c}{\hd{Visual Quality}}
& \multicolumn{3}{c}{\hd{Robot Morphology}}
& \multicolumn{3}{c}{\hd{Robot Kinematics}}
& \hd{Object Dynamics} \\
\cmidrule(lr){2-4}\cmidrule(lr){5-7}\cmidrule(lr){8-10}\cmidrule(l){11-11}
& PSNR\,$\uparrow$ & SSIM\,$\uparrow$ & LPIPS\,$\downarrow$
& Sym.\ IoU\,$\uparrow$ & Boundary $F_1$\,$\uparrow$ & Region LPIPS\,$\downarrow$
& PCK\,$\uparrow$ & Kpt.\ px.\ error\,$\downarrow$ & nDTW\,$\downarrow$
& Trajectory px.\ error\,$\downarrow$ \\
\midrule
\multicolumn{11}{c}{\hd{Seen-pool reference from the multi-view checkpoint}} \\
reference & 27.215 & 0.919 & 0.052 & 0.817 & 0.718 & 0.044 & 0.600 & 11.154 & 0.062 & 9.803 \\
\addlinespace[2pt]
\multicolumn{11}{c}{\hd{Held-out Franka recovery}} \\
$M=0$ & 19.661 & 0.850 & 0.167 & 0.679 & 0.476 & 0.230 & 0.463 & 21.347 & 0.109 & 15.831 \\
$M=1$ & 23.531 & 0.879 & 0.117 & 0.748 & 0.533 & 0.158 & 0.432 & 20.669 & 0.107 & 16.851 \\
$M=2$ & 24.016 & 0.884 & 0.107 & 0.757 & 0.543 & 0.142 & 0.431 & 20.949 & 0.108 & 16.712 \\
$M=3$ & 24.360 & 0.887 & 0.102 & 0.758 & 0.540 & 0.136 & 0.434 & 20.672 & 0.107 & 16.856 \\
\addlinespace[2pt]
\multicolumn{11}{c}{\hd{Held-out Piper recovery}} \\
$M=0$ & 25.778 & 0.908 & 0.090 & 0.777 & 0.662 & 0.128 & 0.592 & 12.482 & 0.083 & 8.285 \\
$M=1$ & 28.188 & 0.923 & 0.061 & 0.815 & 0.738 & 0.093 & 0.617 & 12.981 & 0.084 & 7.859 \\
$M=2$ & 28.497 & 0.927 & 0.055 & 0.815 & 0.749 & 0.084 & 0.622 & 13.044 & 0.081 & 7.685 \\
$M=3$ & 28.801 & 0.929 & 0.053 & 0.817 & 0.751 & 0.080 & 0.612 & 12.555 & 0.081 & 7.787 \\
\bottomrule
\end{tabular*}
\caption{\textbf{Complete few-shot recovery evaluation.} $M=0$ is the
multi-view zero-shot checkpoint from which all few-shot models are resumed.}
\label{tab:fewshot-recovery-full}
\end{table}

\begin{table}[H]
\centering
\scriptsize
\setlength{\tabcolsep}{1.35pt}
\renewcommand{\arraystretch}{1.10}
\begin{tabular*}{\textwidth}{@{\extracolsep{\fill}}l ccc ccc ccc c@{}}
\toprule
\multicolumn{1}{c}{\multirow{2}{*}[-0.5ex]{\hd{Adaptation target}}} & \multicolumn{3}{c}{\hd{Visual Quality}}
& \multicolumn{3}{c}{\hd{Robot Morphology}}
& \multicolumn{3}{c}{\hd{Robot Kinematics}}
& \hd{Object Dynamics} \\
\cmidrule(lr){2-4}\cmidrule(lr){5-7}\cmidrule(lr){8-10}\cmidrule(l){11-11}
& PSNR\,$\uparrow$ & SSIM\,$\uparrow$ & LPIPS\,$\downarrow$
& Sym.\ IoU\,$\uparrow$ & Boundary $F_1$\,$\uparrow$ & Region LPIPS\,$\downarrow$
& PCK\,$\uparrow$ & Kpt.\ px.\ error\,$\downarrow$ & nDTW\,$\downarrow$
& Trajectory px.\ error\,$\downarrow$ \\
\midrule
none (zero-shot UR5) & 30.354 & 0.928 & 0.043 & 0.811 & 0.627 & 0.048 & 0.528 & 16.432 & 0.071 & 13.479 \\
Franka, $M=1$ & 25.304 & 0.905 & 0.073 & 0.799 & 0.612 & 0.083 & 0.474 & 18.119 & 0.078 & 14.693 \\
Franka, $M=2$ & 25.806 & 0.903 & 0.073 & 0.801 & 0.611 & 0.082 & 0.464 & 18.411 & 0.079 & 14.605 \\
Franka, $M=3$ & 26.366 & 0.904 & 0.071 & 0.799 & 0.609 & 0.079 & 0.465 & 18.183 & 0.078 & 14.827 \\
Piper, $M=1$  & 28.557 & 0.909 & 0.057 & 0.825 & 0.664 & 0.062 & 0.553 & 16.053 & 0.070 & 14.166 \\
Piper, $M=2$  & 28.055 & 0.907 & 0.060 & 0.828 & 0.673 & 0.065 & 0.560 & 16.024 & 0.069 & 14.157 \\
Piper, $M=3$  & 28.312 & 0.908 & 0.059 & 0.829 & 0.671 & 0.063 & 0.558 & 15.885 & 0.069 & 14.058 \\
\bottomrule
\end{tabular*}
\caption{\textbf{Complete forgetting evaluation on seen UR5.} Each row scores
the same 500 disjoint UR5 episodes after adapting to the indicated held-out
robot.}
\label{tab:fewshot-forgetting-full}
\end{table}

\FloatBarrier
\subsection{Additional Qualitative Rollouts}
\label{sec:supp-qualitative}
We show additional qualitative rollouts for the action-representation, conditioning, and few-shot-recovery experiments in the following figures.

\begin{figure}[H]
\centering
\includegraphics[width=\columnwidth]{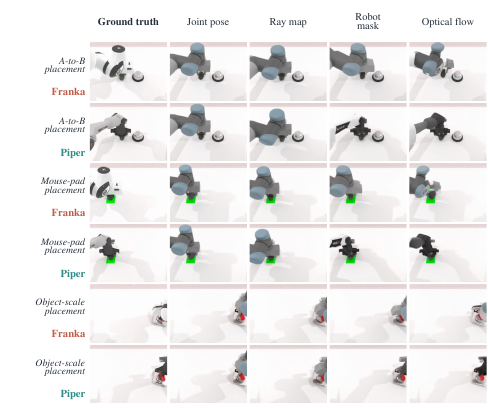}
\caption{\textbf{Action-interface transfer across additional episodes.}
Each task contributes paired Franka and Piper rows.  Columns hold the task,
seed, target robot, and rollout time fixed while varying only the action
representation.  Dense pixel-space controls more consistently retain the
held-out robot than numeric joints across A-to-B, mouse-pad, and object-scale
placement.}
\label{fig:supp-action-rgb}
\end{figure}

\begin{figure}[H]
\centering
\includegraphics[width=\columnwidth]{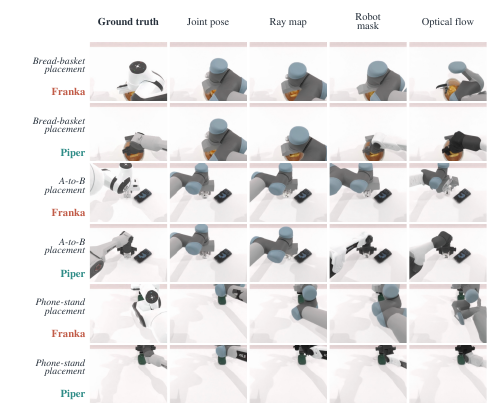}
\caption{\textbf{Additional action-interface rollouts.}
Paired robot rows again isolate the action representation while
holding scene state and rollout time fixed.}
\label{fig:supp-action-rgb-2}
\end{figure}

\begin{figure}[H]
\centering
\includegraphics[width=\columnwidth]{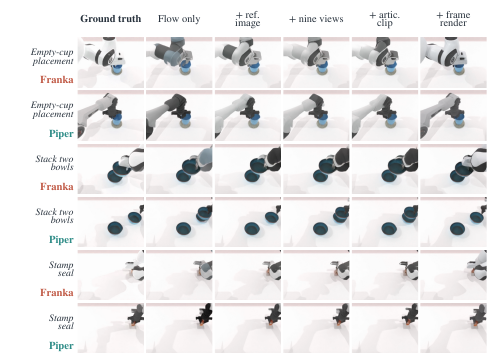}
\caption{\textbf{Robot conditioning across additional episodes.}
Each task again contributes paired Franka and Piper rows.  Reference images,
views, and articulation clips provide task-independent identity cues.}
\label{fig:supp-description-rgb}
\end{figure}

\begin{figure}[H]
\centering
\includegraphics[width=\columnwidth]{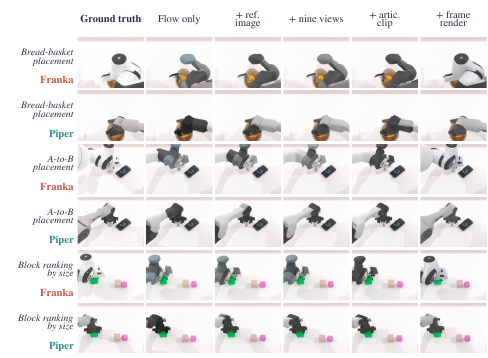}
\caption{\textbf{Additional robot-conditioning rollouts.}
The per-frame condition most directly restores target identity
because its cue is aligned with the current rollout configuration.}
\label{fig:supp-description-rgb-2}
\end{figure}

\FloatBarrier

\end{document}